\documentclass[9pt]{article}
\usepackage{amsmath,graphicx,mlspconf,amssymb}
\usepackage{booktabs,color,nicefrac}
\usepackage{tikz}
\usepackage{hyperref}
\let\OLDthebibliography\thebibliography
\renewcommand\thebibliography[1]{
  \OLDthebibliography{#1}
  \setlength{\parskip}{0pt}
  \setlength{\itemsep}{3.9pt}
}

\newcommand{\tS}{\textsuperscript}

\copyrightnotice{978-1-5090-0746-2/16/\$31.00 {\copyright}2016 IEEE}

\toappear{2016 IEEE International Workshop on Machine Learning for Signal Processing, Sept.\ 13--16, 2016, Salerno, Italy}

\def\x{{\mathbf x}}
\def\y{{\mathbf y}}
\def\X{{\mathbf X}}
\def\Y{{\mathbf Y}}
\def\A{{\mathbf A}}
\def\W{{\mathbf W}}
\def\c{{\mathbf c}}
\def\F{{\mathbf F}}
\def\f{{\mathbf f}}
\def\L{{\cal L}}
\title{A Fully Convolutional Deep Auditory Model for Musical Chord Recognition}
\name{Filip Korzeniowski and Gerhard Widmer
      \thanks{This work is supported by the European Research Council (ERC)
              under the EU's Horizon 2020 Framework Programme (ERC Grant
              Agreement number 670035, project "Con Espressione"). The Tesla
              K40 used for this research was donated by the NVIDIA
              Corporation.}
      }
\address{Johannes Kepler University, Linz, Austria, \\
         Department of Computational Perception}
\begin{document}
%

\maketitle
\begin{abstract}

Chord recognition systems depend on robust feature extraction pipelines. While
these pipelines are traditionally hand-crafted, recent advances in end-to-end
machine learning have begun to inspire researchers to explore data-driven
methods for such tasks. In this paper, we present a chord recognition system
that uses a fully convolutional deep auditory model for feature extraction.
The extracted features are processed by a Conditional Random Field that decodes
the final chord sequence. Both processing stages are trained automatically
and do not require expert knowledge for optimising parameters. We show that the
learned auditory system extracts musically interpretable features, and that the
proposed chord recognition system achieves results on par or better than
state-of-the-art algorithms.
\end{abstract}
\begin{keywords} chord recognition, convolutional neural networks, conditional
random fields \end{keywords}
\section{Introduction} \label{sec:introduction}

Chord Recognition is a long-standing topic of interest in the music information
research (MIR) community. It is concerned with recognising (and transcribing)
chords in audio recordings of music, a labor-intensive task that requires
extensive musical training if done manually. Chords are a highly descriptive
feature of music and useful e.g.\ for creating lead sheets for musicians or as
part of higher-level tasks such as cover song identification.

A chord can be defined as multiple notes perceived simultaneously in harmony.
This does not require the notes to be \emph{played} simultaneously---a melody
or a chord arpeggiation can imply the perception of a chord, even if
intertwined with out-of-chord notes. Through this perceptual
process, the identification of a chord is sometimes subject to interpretation
even among trained experts. This inherent subjectivity is evidenced by
diverse ground-truth annotations for the same songs and discussions about
proper evaluation metrics \cite{humphrey_four_2015}.

Typical chord recognition pipelines comprise three stages: feature extraction,
pattern matching, and chord sequence decoding. Feature extraction transforms
audio signals into representations which emphasise content related to harmony.
Pattern matching assigns chord labels to such representations but works
on single frames or local context only. Chord sequence decoding puts the local
detection into global context by predicting a chord sequence for the complete
audio.

Originally hand-crafted \cite{fujishima_realtime_1999}, all three stages have
seen attempts to be replaced by data-driven methods. For feature extraction,
linear regression \cite{chen_chord_2012}, feed-forward neural networks
\cite{korzeniowski_feature_2016} and convolutional neural networks
\cite{humphrey_learning_2012} were explored; these approaches fit a
transformation from a general time-frequency representation to a manually
defined one that is specifically useful for chord recognition, like chroma
vectors or a ``Tonnetz'' representation. Pattern matching often uses Gaussian
mixture models \cite{cho_improved_2014}, but has seen work on chord
classification directly from a time-frequency representation using
convolutional neural networks \cite{humphrey_rethinking_2012}.
For sequence decoding, hidden Markov models \cite{sheh_chord_2003}, conditional
random fields \cite{burgoyne_cross-validated_2007} and recurrent neural
networks \cite{boulanger-lewandowski_audio_2013} are natural choices; however,
the vast majority of chord recognition systems still rely on hidden Markov
models (HMMs): only one approach used conditional random fields (CRFs) in
combination with simple chroma features for this task
\cite{burgoyne_cross-validated_2007}, with limited success. This warrants
further exploration of this model class, since it has proven to out-perform
HMMs in other domains.

In this paper, we present a novel end-to-end chord recognition system that
combines a fully convolutional neural network (CNN) for feature extraction
with a CRF for chord sequence decoding.
Fully convolutional neural networks replace the stack of dense layers
traditionally used in CNNs for classification with \emph{global average
pooling} (GAP) \cite{lin_network_2013}, which reduces the number of trainable
parameters and improves generalisation. Similarly to
\cite{humphrey_rethinking_2012}, we train the CNN to directly predict chord
labels for each audio frame, but instead of using these predictions directly,
we use the hidden representation computed by the CNN as features for the
subsequent pattern matching and chord sequence decoding stage. We call the
feature-extracting part of the CNN \emph{auditory model.}

For pattern matching and chord sequence decoding, we connect a CRF to the
auditory model. Combining neural networks with CRFs gives
a fully differentiable model that can be learned jointly, as shown in
\cite{peng_conditional_2009,do_neural_2010}.
For the task at hand, however, we found it advantageous to train both parts
separately, both in terms of convergence time and performance.


\section{Feature Extraction}
\label{sec:feature_extraction}

Feature extraction is a two-phase process. First, we convert the signal into a
time-frequency representation in the pre-processing stage. Then, we feed this
representation to a CNN and train it to classify
chords. We take the activations of a hidden layer in the network as high-level
feature representation, which we then use to decode the final chord sequence.

\subsection{Pre-processing}

The first stage of our feature extraction pipeline transforms the input audio
into a time-frequency representation suitable as input to a CNN. As described
in Sec.~\ref{sec:auditory_model}, CNNs consist of fixed-size filters that
capture local structure, which requires the spatial relations to be similarly
distributed in each area of the input. To achieve this, we compute the
magnitude spectrogram of the audio and apply a filterbank with logarithmically
spaced triangular filters. This gives us a time-frequency representation in
which distances between notes (and their harmonics) are equal in all areas of
the input. Finally, we logarithmise the filtered magnitudes to compress the
value range. Mathematically, the resulting time-frequency representation
\(\textbf{L}\) of an audio recording is defined as \[ \mathbf{L} = \log\left(1
+ \mathbf{B}^{\triangle}_{\text{Log}} \left\lvert \mathbf{S}\right\rvert
\right), \] where \(\mathbf{S}\) is the short-time Fourier transform (STFT) of
the audio, and \(\mathbf{B}^{\triangle}_{\text{Log}}\) is the logarithmically
spaced triangular filterbank. To be concise, we will refer to \(\mathbf{L}\) as
\emph{spectrogram} in the remainder of this paper.

We feed the network spectrogram frames with context, i.e.\ the input to the
network is not a single column \(\mathbf{l}_i\) of \(\mathbf{L}\), but a matrix
\(\mathbf{X}_i = \left[\mathbf{l}_{i - C}, \ldots, \mathbf{l}_{i}, \ldots,
\mathbf{l}_{i + C}\right]\), where \(i\) is the index of the target frame, and
\(C\) is the context size.

We chose the parameter values based on our previous study on data-driven
feature extraction for chord recognition \cite{korzeniowski_feature_2016} and a
number of preliminary experiments. We use a frame size of 8192 with a hop size
of 4410 at a sample rate of 44100 Hz for the STFT.
The filterbank comprises 24 filters per octave between
65 Hz and 2100 Hz. The context size \(C = 7\), thus each \(\mathbf{X}_i\)
represents 1.5 sec.\ of audio. Our choice of parameters results in an
input dimensionality of \(\mathbf{X}_i \in \mathbb{R}^{105 \times 15}\).

\subsection{Auditory Model}
\label{sec:auditory_model}

To extract discriminative features from the input, in
\cite{korzeniowski_feature_2016}, we used a simple deep neural network (DNN) to
compute \emph{chromagrams}, concise descriptors of harmonic content. From these
chromagrams, we used a simple classifier to predict chords in a frame-wise
manner. Despite the network being simple conceptually, due to the dense
connections between layers, the model had 1.9 million parameters.

In this paper, we use a CNN for feature extraction. CNNs differ from
traditional deep neural networks by including two additional types of
computational layers: \emph{convolutional layers} compute a 2-dimensional
convolution of their input with a set of fixed-sized, trainable kernels per
feature map, followed by a (usually non-linear) activation function;
\emph{pooling layers} sub-sample the input by aggregating over a local
neighbourhood (e.g.\ the maximum of a $2\times2$ patch). The former can be
re-formulated as a dense layer using a sparse weight matrix with tied weights.
This interpretation indicates the advantages of convolutional layers: fewer
parameters and better generalisation.

CNNs typically consist of convolutional lower layers that act as feature
extractors, followed by fully connected layers for classification. Such layers
are prone to over-fitting and come with a large number of parameters. We thus
follow \cite{lin_network_2013} and use \emph{global average pooling} (GAP) to
replace them. To further prevent over-fitting, we apply
\emph{dropout} \cite{srivastava_dropout_2014},
and use \emph{batch normalisation} \cite{ioffe_batch_2015}
to speed up training convergence.

Table~\ref{tab:convnet} details our model architecture, which consists of 900k
parameters, roughly 50\% of the original DNN. Inspired by the architecture
presented in \cite{simonyan_very_2014}, We opted for multiple lower
convolutional layers with small $3\times3$ kernels, followed by a layer
computing 128 feature maps using $12\times9$ kernels. The intuition is that
these bigger kernels can aggregate harmonic information for the classification
part of the network. We will denote the output of this layer as $\F_i$, the
\emph{features} extracted from input $\X_i$.

\begin{table}[]
\centering
\begin{tabular}{@{}lrcr@{}}
\toprule
\textbf{Layer Type} & \textbf{Parameters} & \textbf{Padding} & \textbf{Output Size}  \\ \midrule
Input               &                     &                  & $105\times15        $ \\
Conv-Rectify        & $32\times3\times3$  & yes              & $32\times105\times15$ \\
Conv-Rectify        & $32\times3\times3$  & yes              & $32\times105\times15$ \\
Conv-Rectify        & $32\times3\times3$  & yes              & $32\times105\times15$ \\
Conv-Rectify        & $32\times3\times3$  & yes              & $32\times105\times15$ \\
Pool-Max            & $2\times1$          &                  & $32\times52\times15 $ \\ \midrule
Conv-Rectify        & $64\times3\times3$  & no               & $64\times50\times13 $ \\
Conv-Rectify        & $64\times3\times3$  & no               & $64\times48\times11 $ \\
Pool-Max            & $2\times1$          &                  & $64\times24\times11 $ \\ \midrule
Conv-Rectify        & $128\times12\times9$& no               & $128\times13\times3 $ \\ \midrule
Conv-Linear         & $25\times1\times1$  & no               & $25\times13\times3  $ \\
Pool-Avg            & $13\times3$         &                  & $25\times1\times1   $ \\
Softmax             &                     &                  & $25                 $ \\ \bottomrule
\end{tabular}
\caption{Proposed CNN architecture. Batch normalisation is performed after each
        convolution layer. Dropout with probability 0.5 is applied at
        horizontal rules in the table. All convolution layers use rectifier
        units \cite{glorot_deep_2011}, except the last, which is linear. The
        bottom three layers represent the GAP, replacing fully connected layers for classification.}
\label{tab:convnet}
\end{table}

We target a reduced chord alphabet in this work (major and minor chords
for 12 semitones) resulting in 24 classes plus a ``no chord'' class. This is a
common restriction used in the literature on chord recognition
\cite{mcvicar_automatic_2014}. The GAP construct thus learns a weighted
average of the 128 feature maps for each of the 25 classes using the $1\times1$
convolution and average pooling layer. Applying the softmax function then
ensures that the output sums to 1 and can be interpreted as a probability
distribution of class labels given the input.

Following \cite{bengio_representation_2013-1}, the activations of the network's
hidden layers can be interpreted as hierarchical feature representations of the
input data. We will thus use $\F_i$ as a feature representation for
the subsequent parts of our chord recognition pipeline.

\subsection{Training and Data Augmentation}

We train the auditory model in a supervised manner using the Adam optimisation
method \cite{kingma_adam_2014} with standard parameters, minimising the
categorical cross-entropy between true targets \(\y_i\) and network output
\(\tilde{\y}_i\). Including a regularisation term, the loss is defined as
\[
        \L = - \frac{1}{D} \sum_{i=1}^{D} \y_i \log(\tilde{\y}_i) + \lambda \left| \mathbf{\theta} \right|_2,
\]
where $D$ is the number of frames in the training data, $\lambda=10^{-7}$ the
$l_2$ regularisation factor, and $\theta$ the network parameters. We
process the training set in mini-batches of size 512, and stop training if
the validation accuracy does not improve for 5 epochs. 

We apply two types of data manipulations to increase the variety of training
data and prevent model over-fitting. Both exploit the fact that the frequency
axis of our input representation is linear in pitch, and thus facilitates the
emulation of pitch-shifting operations. The first operation, as explored in
\cite{humphrey_rethinking_2012}, shifts the spectrogram up or down in discrete
semitone steps by a maximum of 4 semitones. This manipulation does not preserve
the label, which we thus adjust accordingly. The second operation emulates a
slight detuning by shifting the spectrogram by fractions of up to 0.4 of a
semitone. Here, the label remains unchanged. We process each data point in a
mini-batch with randomly selected shift distances. The network thus almost
never sees exactly the same input during training. We found these data
augmenting operations to be crucial to prevent over-fitting.

\section{Chord Sequence Decoding}
\label{sec:chord_sequence_decoding}

Using the predictions of the pattern matching stage directly (in our case, the
predictions of the CNN) often gives good results in terms of frame-wise
accuracy. However, chord sequences obtained this way are often
fragmented. The main purpose of chord sequence decoding is thus to smooth
the reported sequence. 
Here, we use a linear-chain CRF
\cite{lafferty_conditional_2001} to introduce inter-frame dependencies and
find the optimal state sequence using Viterbi decoding.

\subsection{Conditional Random Fields}

Conditional random fields are probabilistic energy-based models for structured
classification. They model the conditional probability distribution
\begin{align}
        p\left(\Y \mid \X\right) = \frac{\exp\left[E\left(\Y, \X\right)\right]} {\sum_{\Y'} \exp\left[E\left(\Y',\X\right)\right]}
\label{eq:cond_prob}
\end{align}
where \(\Y\) is the label vector sequence \(\left[\y_0, \ldots, \y_N\right]\),
and \(\X\) the feature vector sequence of same length. We assume each \(\y_i\)
to be the target label in one-hot encoding. The energy function is defined as
\begin{align}
\begin{split}
E\left(\Y, \X\right) &= \sum_{i=1}^N\left[\y_{n-1}^\top \A \y_n + \y_n^\top \c
                                          + \x_n^\top \W \y_n  \right] \\
                     &\quad + \y_0^\top\pi + \y_N^\top\tau
\end{split}
\label{eq:energy_func}
\end{align}
where \(\A\) models the inter-frame potentials, \(\W\) the frame-input
potentials, \(\c\) the label bias, \(\pi\) the potential of the first label,
and \(\tau\) the potential of the last label. This form of energy function
defines a linear-chain CRF.


From Eq.~\ref{eq:cond_prob} and~\ref{eq:energy_func} follows that
a CRF can be seen as generalised logistic regression. They become
equivalent if we set \(\A\), \(\pi\) and \(\tau\) to 0. Further, logistic
regression is equivalent to a softmax output layer of a neural network. We thus
argue that a CRF whose input is computed by a neural network can be interpreted
as a generalised softmax output layer that allows for dependencies between
individual predictions. This makes CRFs a natural choice for incorporating
dependencies between predictions of neural networks.

\subsection{Model Definition and Training}

Our model has 25 states ($\text{12 semitones} \times \{\text{major}, \text{minor}\}$ and
a ``no-chord'' class).
These states are connected to observed features through the weight matrix $\W$,
which computes a weighted sum of the features for each class.
This corresponds to what the global-average-pooling part of the CNN does.
We will thus use the input to the GAP-part, $\F_i$, averaged for each of the
128 feature maps, as input to the CRF. We can pull the averaging operation
from the last layer to right after the feature-extraction layer, because the
operations in between (linear convolution, batch normalisation) are linear and
no dropout is performed at test-time.

Formally, we will denote the input sequence as $\overline{\F} \in
\mathbb{R}^{128 \times N}$, where each column $\overline{\f}_i$ is the averaged
feature output of the CNN for a given input \(\X_i\). Our CRF thus models
$p\left(\Y \mid \overline{\F}\right)$.

As with the CNN, we train the CRF using Adam, but set a higher learning
rate of 0.01. The mini-batches consist of 32 sequences with a length of 1024
frames (102.3 sec) each. As optimisation criterion, we use the $l_1$-regularized
negative log-likelihood of all sequences in the data set:
\[
        \L = -\frac{1}{S} \sum_{i=1}^S \log p\left(\Y_i \mid \overline{\F}_i\right) + \lambda \left|\xi\right|_1,
\]
where $S$ is the number of sequences in the data set, $\lambda = 10^{-4}$ is
the $l_1$ regularization factor, and $\xi$ are the CRF parameters. We stop
training when validation accuracy does not increase for 5 epochs.

\section{Experiments}
\label{sec:experiments}

\begin{table}
\begin{center}
        \begin{tabular}{@{}rccc@{}}
\toprule
\textbf{} & \textbf{Isophonics} & \textbf{Robbie Williams} & \textbf{RWC} \\ \midrule
CB3       & 82.2                & -                        & -            \\
KO1       & 82.7                & -                        & -            \\
NMSD2     & 82.0                & -                        & -            \\ \midrule
Proposed  & 82.9                & 82.8                     & 82.5         \\ \bottomrule
\end{tabular}

\end{center}
\caption{
Weighted Chord Symbol Recall of major and minor chords achieved
by different algorithms. The results of NMSD2 are statistically significantly
worse than others, according to a Wilcoxon signed-rank test.
Note that train and test data overlaps for CB3, KO1 and NMSD2, while
the results of our method are determined by 8-fold cross-validation.
}
\label{tab:result_table}
\end{table}

We evaluate the proposed system using 8-fold cross-validation on a compound
dataset that comprises the following subsets:
\textbf{Isophonics\footnote{\url{http://isophonics.net/datasets}}:} 180 songs by The Beatles, 19 songs by Queen, and 18 songs by Zweieck,
totalling 10:21 hours of audio.
\textbf{RWC Popular \cite{goto_rwc_2002}:} 100 songs in the style of American and Japanese pop music originally recorded for this data set, totalling 6:46 hours of audio. \textbf{Robbie Williams \cite{di_giorgi_automatic_2013}:} 65 songs by Robbie Williams, totalling 4:30 hours of audio.

As evaluation measure, we compute the Weighted Chord Symbol Recall (WCSR),
often called Weighted Average Overlap Ratio (WAOR), of major and minor chords
as implemented in the ``mir\_eval'' library \cite{raffel_mireval_2014}: \(
        \mathcal{R} = \nicefrac{t_{c}}{t_{a}}\),
where $t_c$ is the total time where the prediction corresponds to the
annotation, and $t_a$ is the total duration of annotations of the respective
chord classes (major and minor chords, in our case).

We compare our results to the three best-performing algorithms in the MIREX
competition in 2013\footnote{\url{http://www.music-ir.org/mirex/wiki/2013:MIREX2013\_Results}}
(no superior algorithm has been submitted to MIREX since then): \textbf{CB3},
based on \cite{cho_improved_2014}; \textbf{KO1},
\cite{khadkevich_time-frequency_2011}; and \textbf{NMSD2},
\cite{ni_using_2012}.

\subsection{Results}

The results presented for the reference algorithms differ from those
found on the MIREX website. This is because of minor differences in the
implementation of the evaluation libraries. To ensure a fairer comparison, we
obtained the predictions of the compared algorithms and ran the same evaluation
code for all approaches. Note however, that for the reference algorithms there
is a known overlap between train and test set, and the obtained results might
be optimistic.

Table~\ref{tab:result_table} shows the results of our method compared to
three state-of-the-art algorithms. We can see that the proposed method performs
slightly better (but not statistically significant), although the
train set of the reference methods overlaps with the test set.

\section{Auditory Model Analysis}
\label{sec:auditory_model_analysis}

\begin{figure}
\centering
\resizebox{0.8\columnwidth}{!}{\includegraphics{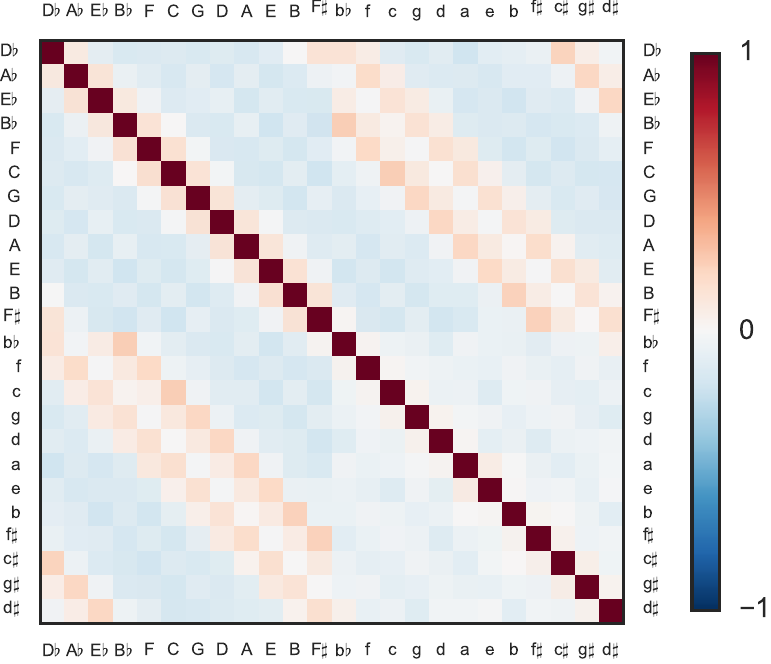}}
\caption{Correlation between weight vectors of chord classes. Rows and columns
         represent chords. Major chords are represented by upper-case letters,
         minor chords by lower-case letters. The order of chords within a
         chord quality is determined by the circle of fifths. We observe that
         weight vectors of chords close in the circle of fifths (such as
         `C', `F', and `G') correlate positively. Same applies to chords that share
         notes (such as `C' and `a', or `C' and `c').}
\label{fig:weight_correlation}
\end{figure}

\begin{figure*}[t!]
\resizebox{\textwidth}{!}{\includegraphics{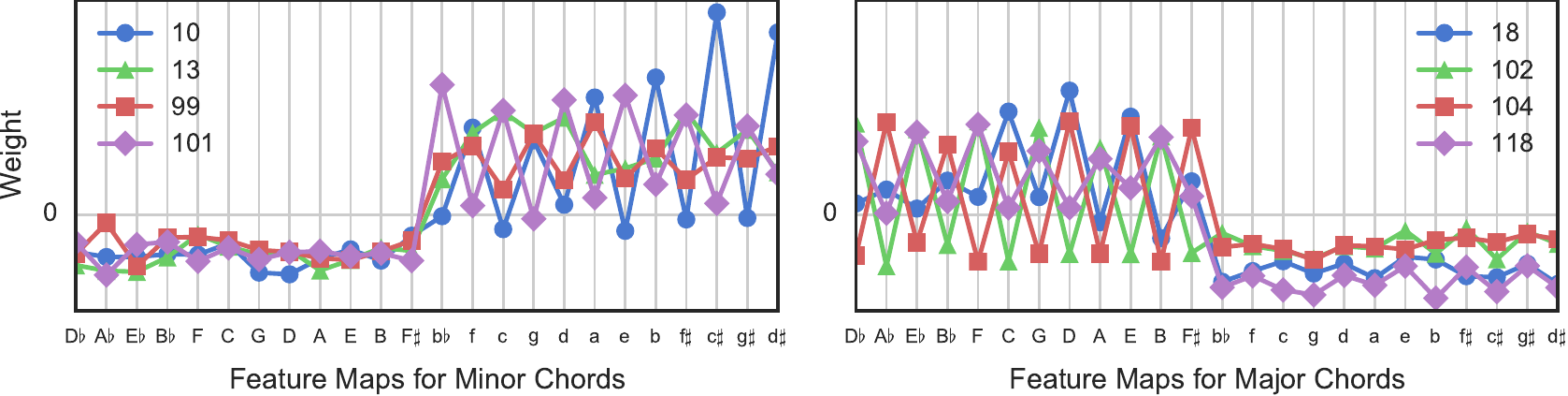}}
\caption{Connection weights of selected feature maps to chord classes. Chord
         classes are ordered according to the circle of fifths, such that
         harmonically close chords are close to each other. In the
         left plot, we selected feature maps that have a high average
         contribution to \emph{minor} chords. In the right plot, those with
         high contribution to \emph{major} chords. Feature maps with high
         average weights to minor chords show negative connections to
         \emph{all} major chords. Within minor chords, we observe that two
         of them (10 and 101) discriminate between chords that are
         harmonically close (zig-zag pattern). We observe a similar pattern
         in the right plot.}
\label{fig:majmin_feature_maps}
\end{figure*}

\begin{figure*}[t!]
\resizebox{\textwidth}{!}{\includegraphics{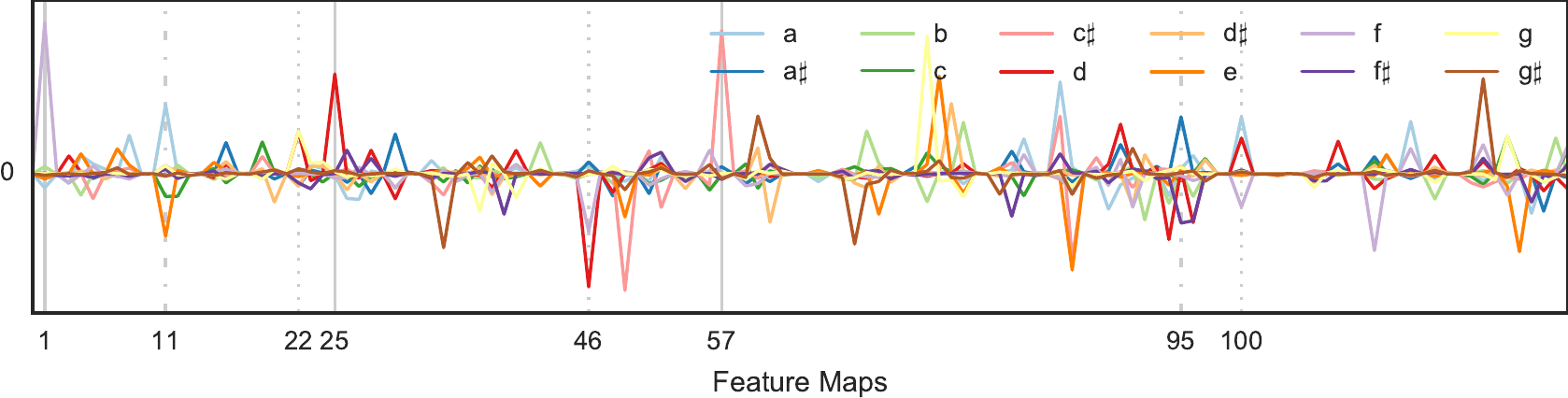}}
\caption{
Contribution of feature maps to pitch classes. Although these results are
noisy, we observe that some feature maps seem to specialise on detecting the
presence (or absence) pitch classes. For example,
feature maps 1, 25, and 57 detect \emph{single} pitch classes; feature maps 22, 46,
and 100 contribute to \emph{pairs} of related pitch classes---perfect fifth between
`g' and `d' in the 22\tS{nd}, minor third between `d' and `f' in the 46\tS{th},
and major third between `a' and `c$\sharp$' in the 85\tS{th} feature map.
Note that the 100\tS{th} feature map also slightly discriminates `d' and `a'
from `f', which would form a d-minor triad together. Other feature maps that
discriminate between pitch classes include the 11\tS{th} (`a' vs.~`e', perfect
fifth) and the 95\tS{th} (`f$\sharp$' vs.`a$\sharp$', major third).
}
\label{fig:pitch_class_feature_maps}
\end{figure*}

Following \cite{lin_network_2013}, the final feature maps of a GAP
network can be interpreted as ``category confidence maps''.  Such
confidence maps will have a high average value if the network is confident that
the input is of the respective category. In our architecture, the average
activation of a confidence map can be expressed as a weighted average over the
(batch-normalised) feature maps of the preceding layer. We thus have 128
weights for each of the 25 categories (chord classes).

We wanted to see whether the penultimate feature maps $\F_i$ can be interpreted
in a musically meaningful way. To this end, we first analysed the similarity of
the weight vectors for each chord class by computing their correlation. The
result is shown in Fig.~\ref{fig:weight_correlation}. We see a systematic
correlation between weight vectors of chords that share notes or are close to
each other in the circle of fifths. The patterns within minor chords are less
clear. This might be because minor chords are under-represented in the data,
and the network could not learn systematic patterns from this limited amount.

Furthermore, we wanted to see if the network learned to distinguish major and
minor modes independently of the root note. To this end, we selected the
four feature maps with the highest connection weights to major and minor
chords respectively and plotted their contribution to each chord class in
Fig.~\ref{fig:majmin_feature_maps}. Here, an interesting pattern emerges:
feature maps with high average weights to minor chords have negative
connections to all major chords. High activations in these feature maps thus
make \emph{all major chords} less likely. However, they tend to be specific on
\emph{which minor chords} they favour. We observe a zig-zag pattern
that discriminates between chords that are next to each other in the circle
of fifths. This means that although the weight vectors of harmonically close
chords correlate, the network learned features to discriminate them.

Finally, we investigated if there are feature maps that indicate the
presence of individual pitch classes. To this end, we multiplied the weight
vectors of \emph{all chords containing a pitch class}, in order to isolate
its influence. For example, when computing the weight vector for pitch class
`c', we multiplied the weight vectors of `C', `F', `A$\flat$', `c', `f', and `a' chords; their only commonality is the presence of the
`c' pitch class. Fig.~\ref{fig:pitch_class_feature_maps} shows the results.
We can observe that some feature maps seem
to specialise in detecting certain pitch classes and intervals, and some
to discriminate between pitch classes.

\section{Conclusion}
\label{sec:conclusion}

We presented a novel method for chord recognition based on a fully
convolutional neural network in combination with a CRF. The method
automatically learns musically interpretable features from the spectrogram, and
performs at least as good as state-of-the-art systems. For future work we
aim at creating a model that distinguishes more chord qualities than major and
minor, independently of the root note of a chord.

\bibliographystyle{IEEEbib}
\bibliography{mlsp2016}

\end{document}